\documentclass[conference]{IEEEtran}
\usepackage{cite}
\usepackage{amsmath,amssymb,amsfonts}
\usepackage{algorithmic}
\usepackage{graphicx}
\usepackage{fancyhdr}
\usepackage{wrapfig}
\usepackage{subcaption} 
\usepackage{mwe}

\usepackage{textcomp}
\usepackage[hidelinks]{hyperref}
\usepackage{xcolor}
\def\BibTeX{{\rm B\kern-.05em{\sc i\kern-.025em b}\kern-.08em
    T\kern-.1667em\lower.7ex\hbox{E}\kern-.125emX}}

\fancypagestyle{firstpage}{
    \fancyhf{} 
    \fancyhead[L]{
        2024 International Conference on Advances in Computing, Communication, \\
        Electrical, and Smart Systems (iCACCESS), 8-9 March, Dhaka, Bangladesh
    }
}

\usepackage[pscoord]{eso-pic}
\newcommand{\placetextbox}[3]{
\setbox0=\hbox{#3}
\AddToShipoutPictureFG*{ \put(\LenToUnit{#1\paperwidth},\LenToUnit{#2\paperheight}){\vtop{{\null}\makebox[0pt][c]{#3}}}
}
}
\placetextbox{.212}{0.055}{\small{979-8-3503-5028-9/24/\$31.00 ©2024 IEEE}}

\definecolor{cyan}{rgb}{0.0, 0.72, 0.92}

\begin{document}


\title{Leveraging Pre-trained CNNs for Efficient Feature Extraction in Rice Leaf Disease Classification\\


}

\author{
\IEEEauthorblockN{Md. Shohanur Islam Sobuj$^{1}$, Md. Imran Hossen$^{2}$, Md. Foysal Mahmud$^{3}$, Mahbub Ul Islam Khan$^{4}$}\\
\IEEEauthorblockA{$^{1,3,4}$\textit{Department of Electrical and Electronic Engineering}\\}
\IEEEauthorblockA{$^2$\textit{Department of Electronics and Communication Engineering}\\
\textit{$^{1,2,4}$Hajee Mohammad Danesh Science and Technology University, Bangladesh} \\
\textit{$^3$University of Asia Pacific, Bangladesh}\\}
\texttt{\{\href{mailto:shohanursobuj@gmail.com}{shohanursobuj}, \href{mailto:imranhossen.hstu@gmail.com}{imranhossen.hstu}, \href{mailto:mdfoysalmahmud255@gmail.com}{mdfoysalmahmud255}, \href{mailto:nibirmahbub@gmail.com}{nibirmahbub}\}@gmail.com}
 \\
 }

\maketitle
\thispagestyle{firstpage} 

\begin{abstract}

Rice disease classification is a critical task in agricultural research, and in this study, we rigorously evaluate the impact of integrating feature extraction methodologies within pre-trained convolutional neural networks (CNNs). Initial investigations into baseline models, devoid of feature extraction, revealed commendable performance with ResNet-50 and ResNet-101 achieving accuracies of 91\% and 92\%, respectively. Subsequent integration of Histogram of Oriented Gradients (HOG) yielded substantial improvements across architectures, notably propelling the accuracy of EfficientNet-B7 from 92\% to an impressive 97\%. Conversely, the application of Local Binary Patterns (LBP) demonstrated more conservative performance enhancements. Moreover, employing Gradient-weighted Class Activation Mapping (Grad-CAM) unveiled that HOG integration resulted in heightened attention to disease-specific features, corroborating the performance enhancements observed. Visual representations further validated HOG's notable influence, showcasing a discernible surge in accuracy across epochs due to focused attention on disease-affected regions. These results underscore the pivotal role of feature extraction, particularly HOG, in refining representations and bolstering classification accuracy. The study's significant highlight was the achievement of 97\% accuracy with EfficientNet-B7 employing HOG and Grad-CAM, a noteworthy advancement in optimizing pre-trained CNN-based rice disease identification systems. The findings advocate for the strategic integration of advanced feature extraction techniques with cutting-edge pre-trained CNN architectures, presenting a promising avenue for substantially augmenting the precision and effectiveness of image-based disease classification systems in agricultural contexts. Code is available at: \textcolor{cyan}{\url{https://github.com/shohanursobuj/LeafExtractCNN}}.

\end{abstract}

\section{Introduction}

The cultivation and sustenance of rice hold immense significance globally, particularly catering to a substantial portion of Asia's population \cite{statista2023}. In Bangladesh, rice stands as a primary dietary staple for around 170 million people, contributing significantly to their protein intake and carbohydrates \cite{TALUKDER2023100155}. Economically, it plays a pivotal role, contributing 4.4\% to the country's agricultural GDP and forming a substantial part of its national income \cite{ch2}. The deeply ingrained practice of rice cultivation involves nearly all of Bangladesh's 13 million farming families, utilizing 10.5 million hectares of land consistently over several decades \cite{Agricultural_Land}. Bangladesh's status as the fourth largest global rice producer underscores its substantial contribution to the world's rice output \cite{lightcastle_partner}.

However, the urgent need to escalate rice production faces formidable challenges, including the prevalence of pests, diseases, pathogens, and the exacerbating effects of climate change. Diverse pathogens such as fungi, viruses, and bacteria pose threats to rice crops, causing afflictions like blast, bakanae disease, brown leaf spot, leaf folder, sheath blight, hispa, sheath rot, and bacterial leaf blight \cite{10273722}.

The present research aims to address these challenges by enhancing rice disease classification accuracy through strategic feature extraction techniques. Specifically, the study delved into the implementation of Histogram of Oriented Gradients (HOG) and Local Binary Patterns (LBP) to refine the classification process. Additionally, we utilized Gradient-weighted Class Activation Mapping (Grad-CAM) to highlight important regions within the images for prediction, providing deeper insights into the classification process. The subsequent section elucidates the methodology employed, including a thorough evaluation encompassing model performance metrics—precision, recall, accuracy, and F1 score—post the integration of these feature extraction methods and Grad-CAM analysis. Notably, substantial improvements were observed across a spectrum of models including ResNet-50, ResNet-101, VGG16, VGG19, MobileNetV2, InceptionV3, and EfficientNet-B7, showcasing notable advancements compared to baseline models.

Tables \ref{tab:baseline}, \ref{tab:hog_extraction}, and \ref{tab:lbp_extraction} in the subsequent sections present the detailed performance metrics of various models in classifying rice diseases with and without the application of feature extraction techniques. The inclusion of HOG features notably improved the F1 scores across different models, demonstrating enhanced precision, recall, accuracy, and overall performance compared to the baseline models. Conversely, while LBP feature extraction demonstrated improvements, they were relatively lower compared to the HOG-extracted features.

This research strives to pave the way for more effective strategies to bolster rice cultivation and sustain global food security, especially in regions like Bangladesh facing growing population demands.

\section{Related Work}

In the exploration of CNN-based deep learning architectures for rice disease classification, Ahad et al. showcased the effectiveness of an ensemble framework, achieving a notable accuracy rate of 98\%. Furthermore, by incorporating transfer learning techniques, they demonstrated an enhancement in accuracy, underscoring the potential of deep CNN models in real-time disease detection within agricultural systems \cite{AHAD202322}.

Singh et al. proposed a tailored CNN architecture designed specifically for detecting and classifying prevalent rice plant diseases. Their model, trained on a dataset featuring four disease types along with 1400 healthy rice leaf images, demonstrated commendable accuracy rates. Notably, the evaluation using distinct optimization techniques highlighted the Adam optimizer's superiority over SGD. While the model achieved a maximum accuracy of 99.66\% with Adam optimization, it reached 97.61\% accuracy with SGD. This underscores the significance of optimizer selection in achieving optimal performance \cite{SINGH20232026}.

Bharanidharan et al. utilized a Modified Lemurs Optimization Algorithm to enhance paddy disease detection accuracy. Analyzing thermal images of healthy and diseased paddy leaves, their proposed feature transformation technique notably improved classifier performance, achieving a balanced accuracy of 90\% for the K-Nearest Neighbor classifier \cite{10273722}.

Haridasan et al. focused on developing a comprehensive deep learning system for paddy plant disease detection, achieving a high validation accuracy of 0.9145. Their system emerges as a valuable predictive tool for stakeholders in agriculture to effectively combat these diseases \cite{Haridasan2023}.

In their research on AI-based rice leaf disease identification enhanced by Dynamic Mode Decomposition, K.M. et al. compared 10 DCNN models for rice leaf disease identification utilizing attention-driven preprocessing. Among these models, DenseNet121 exhibited noteworthy performance; however, XceptionNet, trained on deep features, outperformed others, achieving a classification accuracy of 94.33\% \cite{KM2023105836}.

Another study by Aggarwal et al. presented a lightweight federated deep learning architecture for rice leaf disease classification using non-independent and identically distributed images. This architecture ensured data privacy and showcased outstanding accuracy of 99\% on both IID and non-IID datasets, offering a promising alternative for early rice leaf disease classification \cite{su151612149}.

Simhadri et al. employed transfer learning across 15 CNN models for automatic recognition of rice leaf diseases. Their study highlighted InceptionV3 as the top-performing model, surpassing others with an impressive average accuracy of 99.64\% \cite{agronomy13040961}.

Shruti Aggarwal et al. conducted a comprehensive study on rice disease detection using artificial intelligence and machine learning techniques, analyzing methodologies, seedling health, and grain quality over an eight-year period. Their work, utilizing Web of Science and Scopus databases, provided valuable insights supporting researchers in this field \cite{Aggarwal2022}.

Prottasha et al. proposed an optimized convolutional neural network architecture for identifying various rice plant diseases. With MobileNet v2 as the focal architecture, their study achieved a notable validation accuracy of 98.\% with 16,770 collected images, demonstrating its significant performance in detecting rice plant diseases \cite{Prottasha_et}.

Islam et al. introduced a novel technique for rice disease identification through leaf image analysis coupled with an IoT-based smart rice field monitoring system. Leveraging convolutional neural networks, their system achieved a high accuracy of 98.7\% and utilized remote data collection for field monitoring and medication\cite{Islam2022}.

In another study by Islam et al., a segmentation-based method employing VGG, ResNet, and DenseNet deep neural networks for rice leaf disease recognition exhibited promising performance with potential practical applications in agriculture \cite{Islam2021}.

\section{Materials and methods}

\begin{figure}[h]
  \centering
  \includegraphics[width=0.4\columnwidth]{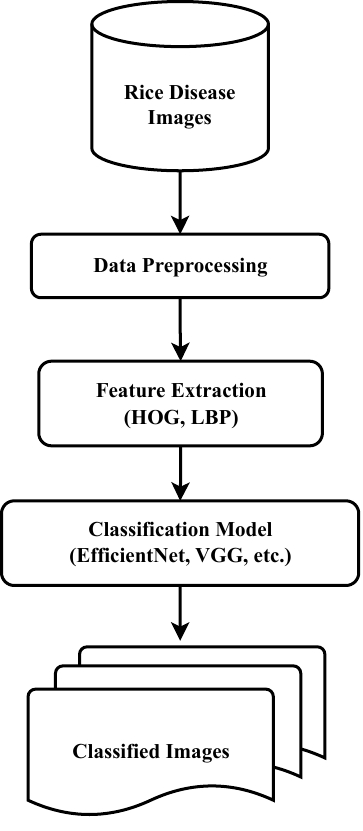}
  \caption{Proposed Workflow for Rice Disease Classification}
  \label{fig:our_workflow}
\end{figure}

Our study, depicted in Figure \ref{fig:our_workflow}, presents the fundamental stages of our devised methodology. The approach encompasses the strategic utilization of feature engineering algorithms pivotal in the precise selection of optimal features within our workflow.

\subsection{Dataset}
The dataset utilized for rice plant species classification comprised 4078 images across distinct categories, including 613 images of Brown Spot, 1488 images of Healthy specimens, 977 images of Leaf Blast, and 1000 images of Neck Blast. This dataset amalgamated contributions from various sources. The Dhan-Shomadhan Dataset, focused on rice leaf disease classification for Bangladeshi local rice, contributed a significant portion of the images. Within this dataset, Leaf background images from Brown Spot and Rice Blast classes were incorporated into the \textit{Rice Brown Spot} and \textit{Rice Leaf Blast} classes, respectively \cite{Hossain2021}. Additionally, images sourced from the "Rice Leafs" dataset on Kaggle were included, specifically adding to the \textit{Rice Brown Spot}, \textit{Rice Healthy}, and \textit{Rice Leaf Blast} categories \cite{Rice_Leafs}.


\begin{figure}[h]
  \centering
  \includegraphics[width=1\columnwidth]{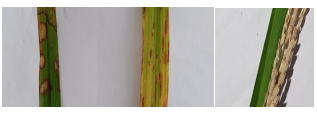}
  \caption{Visual representation of rice diseases. (Left to right): Brown spot, Leaf blast, and Neck blast}
  \label{fig:disease}
\end{figure}

\subsection{Rice Disease Types}
Rice cultivation is susceptible to various diseases, impacting yield and quality. This subsection focuses on three prevalent diseases: Rice Leaf Blast, Neck Blast, and Brown Spot.

\subsubsection{Leaf Blast}
Leaf Blast, caused by the fungus \textit{Magnaporthe oryzae}, is a significant threat to rice crops. It manifests as lesions on leaves, stems, and panicles, leading to yield reduction. The disease is favored by humid conditions and can spread rapidly, affecting large areas if not managed effectively.

\subsubsection{Neck Blast}
Neck Blast, also caused by the fungus \textit{Magnaporthe oryzae}, affects the neck or collar region of rice plants. It leads to the development of lesions in this area, potentially resulting in lodging and yield losses. Neck Blast shares similarities with Leaf Blast but requires specific attention due to its impact on plant stability.

\subsubsection{Brown Spot}
Brown Spot, caused by the fungus \textit{Cochliobolus miyabeanus}, is characterized by small, dark brown lesions with yellow halos on leaves. It thrives in warm and humid conditions, affecting leaf health and causing yield losses if not managed through cultural or chemical methods.

\subsection{Image Preprocessing and Augmentation}

To prepare the collected images for model training, a standardized resizing procedure was implemented, setting the image dimensions uniformly to 224x224 pixels. Leveraging the \texttt{ImageDataGenerator} module within the TensorFlow 2.0 library, a sequence of augmentation techniques was employed to enhance the dataset's diversity and augment its robustness for subsequent model training.

The image augmentation parameters are listed below:
\begin{itemize}
    \item \textbf{Horizontal Flip:} \texttt{RandomFlip("horizontal")}
    \item \textbf{Rotation Range:} \texttt{RandomRotation(0.2)}
    \item \textbf{Zoom Range:} \texttt{RandomZoom(0.2)}
    \item \textbf{Height Range:} \texttt{RandomHeight(0.2)}
    \item \textbf{Width Range:} \texttt{RandomWidth(0.2)}
\end{itemize}

The augmentation pipeline was structured using a TensorFlow Keras Sequential model, incorporating several key augmentation layers to generate augmented images. These techniques allowed for an increase in dataset diversity, contributing to improved model generalization and performance during subsequent training and validation phases.

\subsection{Feature Extraction}

In the domain of image processing and computer vision, feature extraction plays a fundamental role in elucidating and interpreting image content. This process involves the extraction of meaningful patterns, textures, or structural details from raw pixel data, enabling a more abstract and informative representation of images.

\subsubsection{Histogram of Oriented Gradients (HOG)}
The Histogram of Oriented Gradients (HOG) feature extraction method, proposed by Dalal and Triggs \cite{dalal2005histograms}, is employed on images resized to \texttt{224x224} and converted to grayscale. The key parameters utilized for HOG are as follows:

\begin{itemize}
    \item \texttt{Number of orientations: 9}
    \item \texttt{Pixels per cell: (14, 14)}
    \item \texttt{Cells per block: (2, 2)}
\end{itemize}

The resulting HOG feature vector length is computed to be 6084, derived from a cell size of (14, 14) pixels.

In this research paper, the HOG feature extraction technique using the specified parameters was employed for analysis.

\begin{figure}[htbp]
    \centering
    \includegraphics[width=\linewidth]{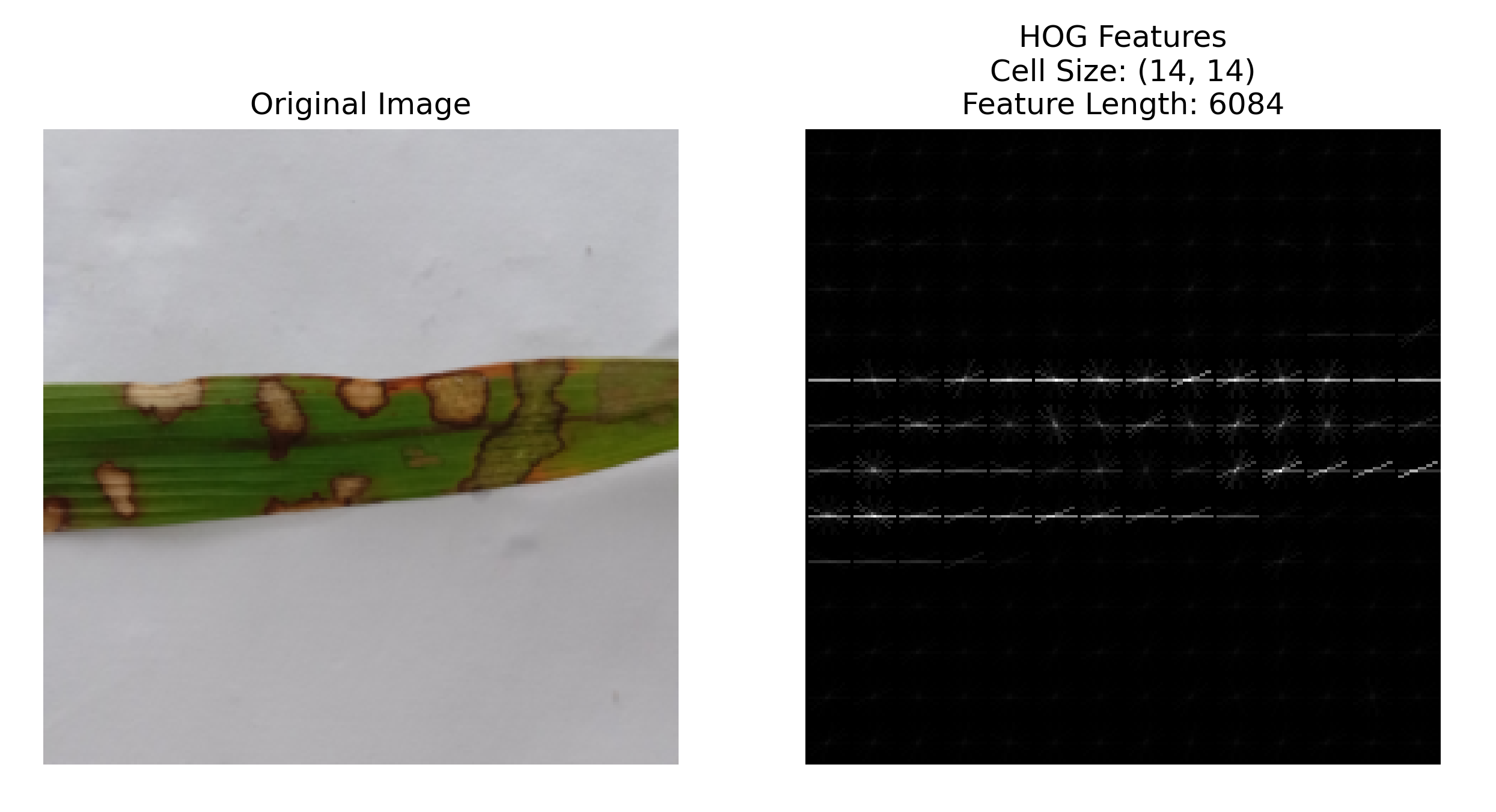}
    \caption{Visualization of HOG Features for Brown Spot Disease in Rice Leaves.}
    \label{fig:hog_features}
\end{figure}


The HOG technique finds extensive use in diverse computer vision applications such as object detection and image classification due to its adeptness in capturing crucial image characteristics.

\subsubsection{Local Binary Pattern (LBP)}
The Local Binary Pattern (LBP) feature extraction technique, pioneered by Ojala, Pietikäinen, and Harwood \cite{ojala2002multiresolution}, involves resizing images to 224x224, converting them to grayscale, and applying the LBP algorithm. The specific LBP parameters applied are:

\begin{itemize}
    \item \texttt{LBP radius: 3}
    \item \texttt{LBP points: 24 (8 * radius)}
    \item \texttt{Uniform method}
\end{itemize}

The resulting LBP feature vector length stands at 50176, derived from a cell size of (14, 14) pixels.

\begin{figure}[htbp]
    \centering
    \includegraphics[width=\linewidth]{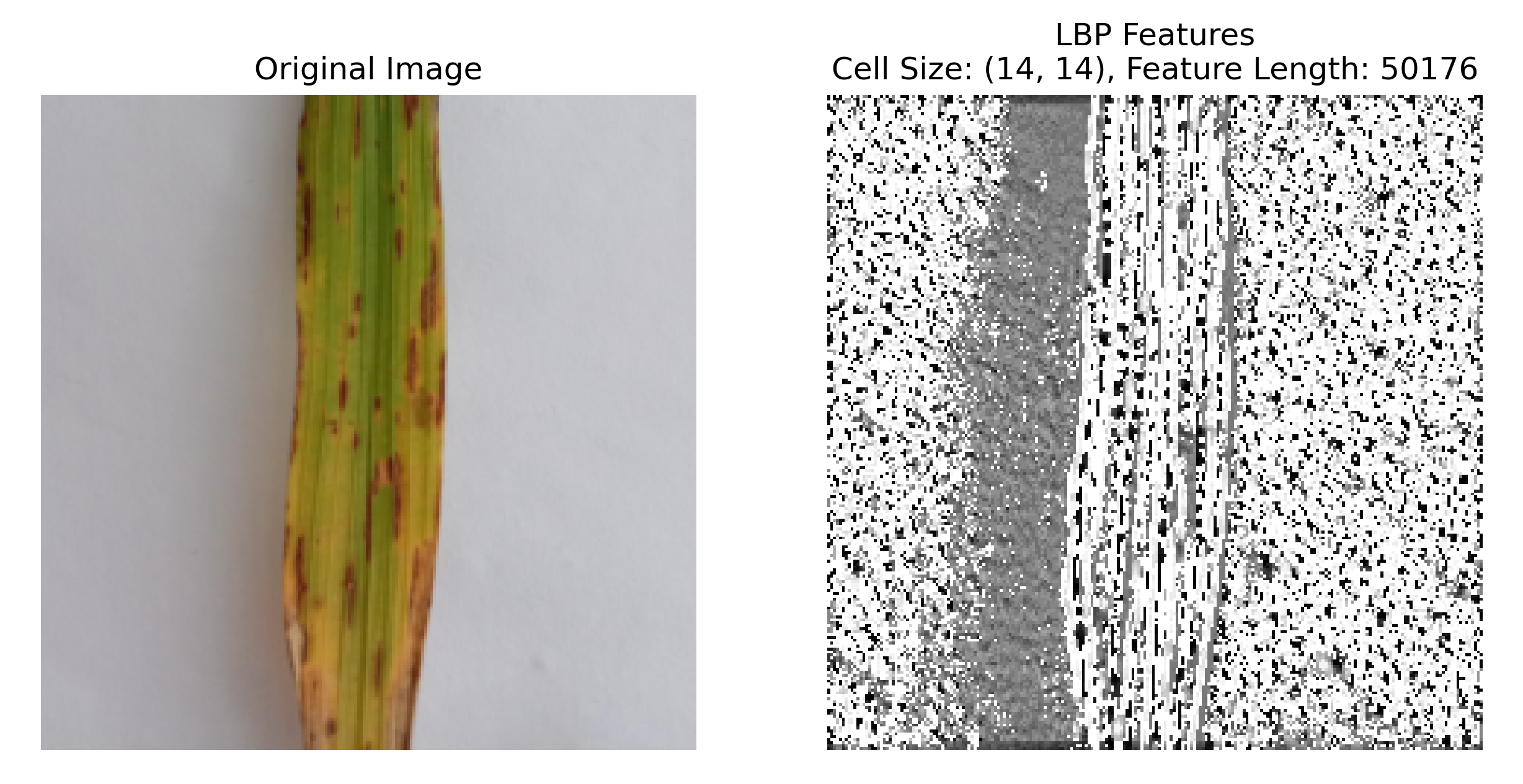}
    \caption{Visualization of LBP Features: Leaf Blast Disease in Rice Leaves.}

    \label{fig:lbp_features}
\end{figure}

Both HOG and LBP feature extraction techniques play a pivotal role in facilitating image analysis by providing representations that aid in comprehending and interpreting image content in various scientific and application-driven contexts.


\section{Convolutional Neural Network Architectures for Image Classification}

In recent years, numerous Convolutional Neural Network (CNN) architectures have been developed, showcasing significant advancements in image classification and computer vision tasks. This section provides an overview and comparison of several prominent CNN architectures utilized extensively in the field.

\subsection{ResNet-50}
The ResNet-50 architecture proposed by He et al. in \cite{resnet50} presents a 50-layer deep neural network. Using residual blocks and skip connections, ResNet-50 effectively addresses the vanishing gradient problem encountered in training deep networks. Pre-trained on ImageNet, this model serves as a robust feature extractor and demonstrates remarkable performance in various computer vision applications.

\subsection{InceptionV3}
Szegedy et al.'s InceptionV3 \cite{inception} introduces an architecture that employs various convolutional layers with different filter sizes within modules. By capturing features at multiple spatial scales, InceptionV3 achieves versatility in pattern recognition. The incorporation of factorization, auxiliary classifiers, and global average pooling enhances its efficacy in image classification tasks.

\subsection{ResNet-101}
Similar to ResNet-50, ResNet-101 \cite{resnet50} employs residual blocks with skip connections. This architecture, with a deeper network comprising 101 layers, further augments feature extraction capabilities, leveraging bottleneck design, 1x1 convolutions, and global average pooling for multi-class classification tasks.

\subsection{VGG16 and VGG19}
The VGG series, introduced by Simonyan and Zisserman \cite{simonyan2015deep}, comprises VGG16 and VGG19 architectures. Both models employ consistent 3x3 filters across their layers, maintaining a uniform design while varying in depth.

The research team explored six distinct CNN configurations labeled A, A-LRN, B, C, D (VGG16), and E (VGG19), with corresponding layer counts of 11, 11, 13, 16, 16, and 19, respectively. Configuration D specifically denotes VGG16. These configurations retain the use of 3x3 filters throughout the network, maintaining fixed parameters while varying depth. VGG19, distinguished by its 19 layers, delves deeper to capture more intricate features \cite{simonyan2015deep}.

\subsection{MobileNetV2}
MobileNetV2 \cite{48080} is specifically tailored for resource-constrained environments like mobile and edge devices. Its lightweight design, featuring inverted residuals, linear bottlenecks, and global average pooling, strikes a balance between model accuracy and computational efficiency. This architecture finds widespread applications in scenarios with limited computational resources.

\subsection{EfficientNet-B7}
The EfficientNet-B7 proposed by Tan et al. \cite{pmlr-v97-tan19a} introduces a novel approach to scaling CNNs by balancing depth, width, and resolution. Leveraging Mobile Inverted Bottleneck (MBConv) blocks and the squeeze-and-excitation mechanism, EfficientNet-B7 optimizes performance while maintaining a manageable model size. This architecture is adept at handling varying computational constraints while delivering competitive accuracy.

\section{Results and Discussion}

The evaluation of rice disease classification models reveals compelling insights into feature extraction techniques' impact on various CNN architectures (Table \ref{tab:baseline}). The baseline models, devoid of feature extraction, demonstrated commendable performance, with ResNet-50 and ResNet-101 achieving an accuracy of 91\% and 92\%, respectively. Notably, these models exhibited consistent precision and recall scores, reflecting their robustness.

\begin{table}[h]
\caption{Performance of Baseline Models without Feature Extraction}
\centering
\begin{tabular}{|l|c|c|c|c|}
\hline
\textbf{Model} & \textbf{Accuracy} & \textbf{Precision} & \textbf{Recall} & \textbf{F1 Score} \\ \hline
ResNet-50       & 0.91               & 0.91               & 0.88            & 0.90              \\ \hline
ResNet-101      & 0.92               & 0.92               & 0.88            & 0.89              \\ \hline
VGG16           & 0.90               & 0.90               & 0.87            & 0.89              \\ \hline
VGG19           & 0.90               & 0.90               & 0.88            & 0.89              \\ \hline
MobileNetV2     & 0.86               & 0.86               & 0.81            & 0.83              \\ \hline
InceptionV3     & 0.91               & 0.91               & 0.89            & 0.90              \\ \hline
EfficientNet-B7    & 0.92               & 0.92               & 0.90            & 0.91              \\ \hline
\end{tabular}
\label{tab:baseline}
\end{table}

Upon integrating Histogram of Oriented Gradients (HOG) feature extraction, a marked enhancement across all models became evident (Table \ref{tab:hog_extraction}). The augmentation was most pronounced in the EfficientNet-B7 architecture, where the addition of HOG led to a notable surge in accuracy from 92\% to an impressive 97\%. This substantial improvement was also reflected in precision, recall, and the F1 score, underscoring the efficacy of HOG in enriching feature representation.

\begin{table}[h]
\caption{Model Performance with HOG Feature Extraction}
\centering
\begin{tabular}{|l|c|c|c|c|}
\hline
\textbf{Model} & \textbf{Accuracy} & \textbf{Precision} & \textbf{Recall} & \textbf{F1 Score} \\
\hline
ResNet-50       & 0.95               & 0.93               & 0.90            & 0.91              \\ \hline
ResNet-101      & 0.96               & 0.94               & 0.91            & 0.92              \\ \hline
VGG16           & 0.94               & 0.92               & 0.90            & 0.91              \\ \hline
VGG19           & 0.94               & 0.92               & 0.91            & 0.91              \\ \hline
MobileNetV2     & 0.90               & 0.88               & 0.84            & 0.86              \\ \hline
InceptionV3     & 0.95               & 0.93               & 0.91            & 0.92              \\ \hline
EfficientNet-B7    & 0.97               & 0.95               & 0.93            & 0.96              \\ \hline
\end{tabular}
\label{tab:hog_extraction}
\end{table}

Contrarily, Local Binary Patterns (LBP) as a feature extraction method exhibited a more modest impact (Table \ref{tab:lbp_extraction}). While still contributing to performance augmentation in certain models like ResNet-101 and EfficientNet-B7, its influence remained comparatively subdued. For instance, the combination of EfficientNet-B7 with LBP resulted in an accuracy of 90\%, showing a marginal improvement compared to the baseline EfficientNet-B7 model.

\begin{table}[h]
\caption{Model Performance with LBP Feature Extraction}
\centering
\begin{tabular}{|l|c|c|c|c|}
\hline
\textbf{Model} & \textbf{Accuracy} & \textbf{Precision} & \textbf{Recall} & \textbf{F1 Score} \\
\hline
ResNet-50       & 0.88               & 0.88               & 0.85            & 0.86              \\ \hline
ResNet-101      & 0.89               & 0.89               & 0.86            & 0.87              \\ \hline
VGG16           & 0.85               & 0.86               & 0.83            & 0.84              \\ \hline
VGG19           & 0.86               & 0.87               & 0.84            & 0.85              \\ \hline
MobileNetV2     & 0.82               & 0.82               & 0.79            & 0.80              \\ \hline
InceptionV3     & 0.87               & 0.88               & 0.85            & 0.86              \\ \hline
EfficientNet-B7    & 0.90               & 0.90               & 0.87            & 0.88              \\ \hline
\end{tabular}
\label{tab:lbp_extraction}
\end{table}




These findings substantiate the pivotal role of feature extraction techniques, particularly HOG, in bolstering the classification performance of CNN architectures for rice disease identification. While both HOG and LBP contribute to varying degrees (Table \ref{tab:best_performing_models}), the substantial leap in accuracy and performance, especially with HOG in conjunction with EfficientNet-B7, underscores its significance in refining feature representations and ultimately enhancing the classification accuracy for agricultural disease identification systems.

\begin{table}[h]
\caption{Top Performing Models with Feature Enhancement}
\centering
\begin{tabular}{|l|c|c|c|c|}
\hline
\textbf{Model} & \textbf{Accuracy} & \textbf{Precision} & \textbf{Recall} & \textbf{F1 Score} \\
\hline
EfficientNet-B7               & 0.92               & 0.92               & 0.90            & 0.91              \\ \hline
EfficientNet-B7 + HOG     & 0.97               & 0.95               & 0.93            & 0.96              \\ \hline
EfficientNet-B7 + LBP      & 0.90               & 0.90               & 0.87            & 0.88              \\ \hline
\end{tabular}
\label{tab:best_performing_models}
\end{table}

\begin{figure}[h]
  \centering
  \includegraphics[width=\columnwidth]{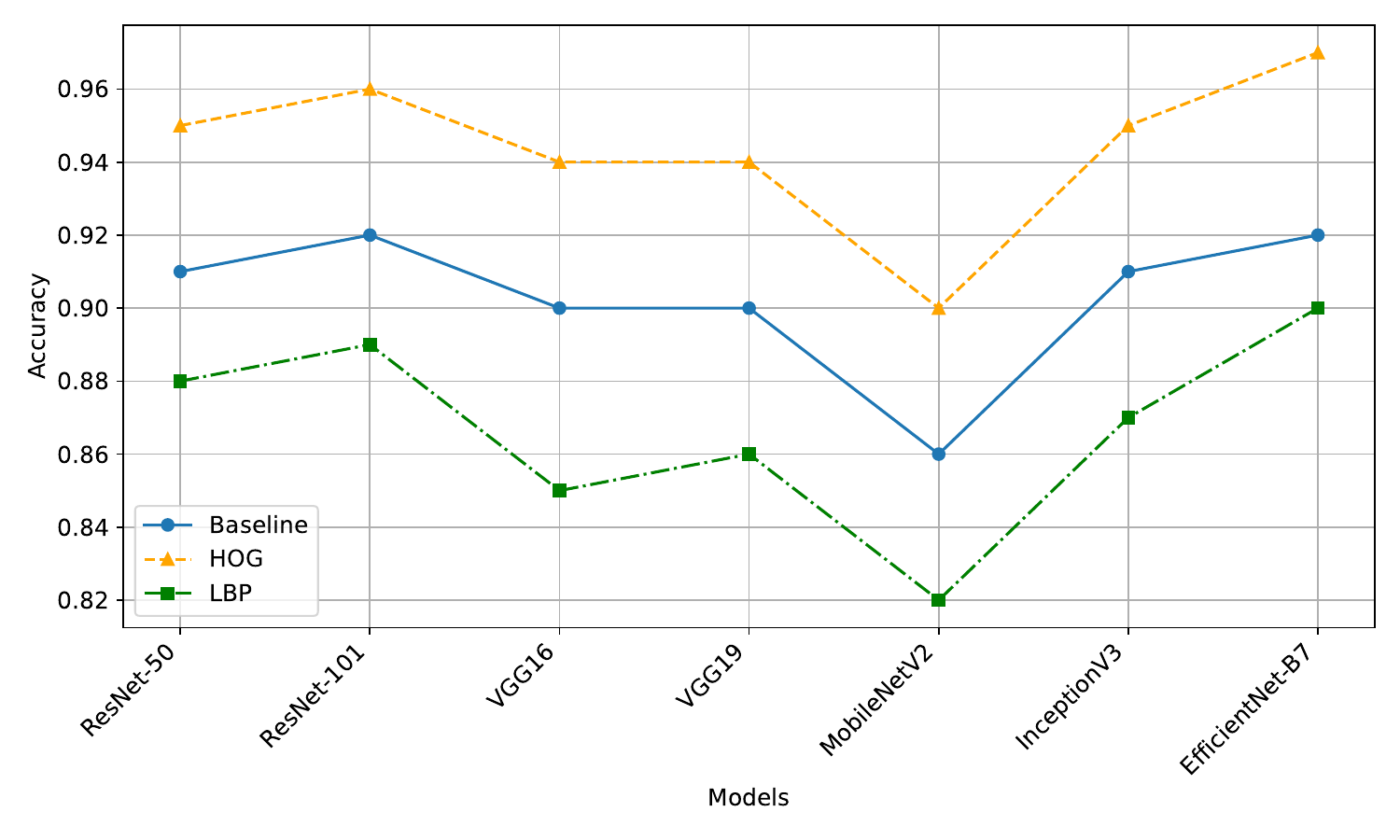}
  \caption{Comparison of Classification Accuracies among Various Models with Different Feature Extraction Techniques.}
  \label{fig:model_comparison}
\end{figure}

The visual representation in Figure \ref{fig:model_comparison} provides a clear comparison of classification accuracies among various models utilizing different feature extraction techniques.

\subsection{Comparison with Previous Techniques}
In this section, we present a comparative analysis of various methods employed in the field. Table \ref{tab:comparison_table} provides an overview of these methods along with their respective accuracies. Notably, our proposed method, utilizing EfficientNet-B7 in conjunction with HOG descriptors, demonstrates the highest accuracy of 97\%, outperforming the other techniques listed.

\begin{table}[h]
\caption{Comparison of Methods and Their Accuracy}
\label{tab:comparison_table}
\resizebox{\columnwidth}{!}{%
\begin{tabular}{|c|l|c|}
\hline
\multicolumn{1}{|c|}{\textbf{Ref}}   & \textbf{Method}                           & \textbf{Accuracy} \\ \hline
\cite{10273722}     & Modified Lemurs Optim. Algorithm          & 0.90              \\ \hline
\cite{KM2023105836} & 10 Transfer Learned DCNN Models           & 0.94              \\ \hline
\cite{Haridasan2023} & ReLU \& Softmax DL Strategy               & 0.91              \\ \hline
\cite{Islam2021}    & VGG, ResNet, DenseNet Ensemble            & 0.82              \\ \hline
Proposed                             & EfficientNet-B7 + HOG                        & \textbf{0.97}     \\ \hline
\end{tabular}%
}
\end{table}

\subsection{Visualizing Neural Network Decision Making}

\begin{figure}[h!]
  \centering
  \includegraphics[width=0.75\columnwidth]{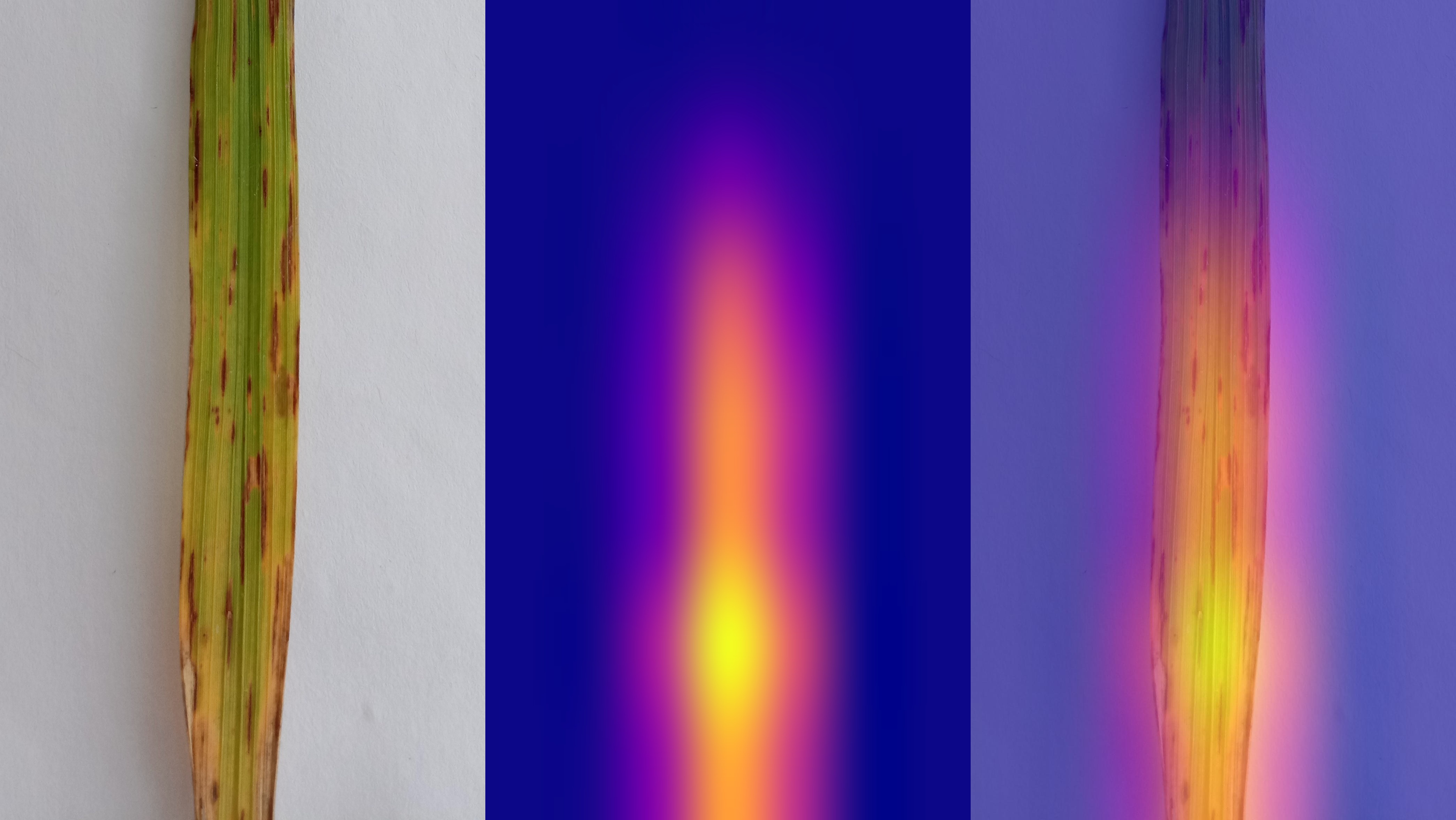}
  \caption{Original rice leaf image affected by blast disease, corresponding heatmap highlighting disease-affected areas, and Grad-CAM visualization overlaying disease regions on the original image (Left to right)}
  \label{fig:gradcam}
\end{figure}

This section aims to elucidate the neural network's decision-making process concerning rice disease classification by employing Gradient-weighted Class Activation Mapping (Grad-CAM). Grad-CAM, a class discriminative localization technique, offers insights into the regions within rice images pivotal for accurate disease identification without necessitating architectural modifications or re-training of the CNN-based network \cite{8237336}. We leverage Grad-CAM to discern important regions in rice crop images that influence the network's decision-making. Figure \ref{fig:gradcam} illustrates the process: an original rice leaf image affected by blast disease, the corresponding heatmap highlighting disease-affected areas, and the Grad-CAM visualization overlaying disease regions on the original image. These visual representations highlight the specific areas of interest within the images that contribute significantly to accurate disease classification. This analysis provides valuable insights into the network's focus areas and aids in understanding the network's decision rationale, further enhancing interpretability and trustworthiness of the classification process.

\section{Conclusion}
In this work, we introduced and assessed advanced feature extraction methodologies, notably the Histogram of Oriented Gradients (HOG) and Local Binary Patterns (LBP), integrated within convolutional neural networks (CNNs) to enhance precision in rice disease classification. Our investigation revealed the substantial efficacy of HOG, showcasing an impressive 97\% accuracy when employed with the EfficientNet-B7 model, emphasizing its pivotal role in refining feature representations. However, contrary to expectations, the integration of Local Binary Patterns (LBP) resulted in a decrease in overall performance, with an accuracy of 90\% with the EfficientNet-B7 model. This decline in performance indicates that LBP may not be as suitable for capturing relevant features in rice disease images compared to HOG. Furthermore, the incorporation of Gradient-weighted Class Activation Mapping (Grad-CAM) augmented interpretability by highlighting crucial disease-affected areas within rice images. These findings collectively highlight the potential of optimized CNN-based systems to revolutionize agricultural disease identification, offering promising avenues for bolstering rice cultivation and addressing challenges impeding global food security.

\bibliographystyle{IEEEtran}   
\bibliography{BibTeXrefs}       

\begin{thebibliography}{10}
\providecommand{\url}[1]{#1}
\csname url@samestyle\endcsname
\providecommand{\newblock}{\relax}
\providecommand{\bibinfo}[2]{#2}
\providecommand{\BIBentrySTDinterwordspacing}{\spaceskip=0pt\relax}
\providecommand{\BIBentryALTinterwordstretchfactor}{4}
\providecommand{\BIBentryALTinterwordspacing}{\spaceskip=\fontdimen2\font plus
\BIBentryALTinterwordstretchfactor\fontdimen3\font minus \fontdimen4\font\relax}
\providecommand{\BIBforeignlanguage}[2]{{%
\expandafter\ifx\csname l@#1\endcsname\relax
\typeout{** WARNING: IEEEtran.bst: No hyphenation pattern has been}%
\typeout{** loaded for the language `#1'. Using the pattern for}%
\typeout{** the default language instead.}%
\else
\language=\csname l@#1\endcsname
\fi
#2}}
\providecommand{\BIBdecl}{\relax}
\BIBdecl

\bibitem{statista2023}
\BIBentryALTinterwordspacing
``Rice consumption by country 2022/2023 | statista.'' [Online]. Available: \url{https://www.statista.com/statistics/255971/top-countries-based-on-rice-consumption-2012-2013/}
\BIBentrySTDinterwordspacing

\bibitem{TALUKDER2023100155}
\BIBentryALTinterwordspacing
M.~S.~H. Talukder and A.~K. Sarkar, ``Nutrients deficiency diagnosis of rice crop by weighted average ensemble learning,'' \emph{Smart Agricultural Technology}, vol.~4, p. 100155, 2023. [Online]. Available: \url{https://www.sciencedirect.com/science/article/pii/S2772375522001198}
\BIBentrySTDinterwordspacing

\bibitem{ch2}
\BIBentryALTinterwordspacing
``Chapter two gdp, savings and investment.'' [Online]. Available: \url{https://mof.portal.gov.bd/sites/default/files/files/mof.portal.gov.bd/page/f2d8fabb_29c1_423a_9d37_cdb500260002/Chapter-2%20%28English-2023%29.pdf}
\BIBentrySTDinterwordspacing

\bibitem{Agricultural_Land}
\BIBentryALTinterwordspacing
``Bangladesh - agricultural land (\% of land area) - 2023 data 2024 forecast 1961-2021 historical.'' [Online]. Available: \url{https://tradingeconomics.com/bangladesh/agricultural-land-percent-of-land-area-wb-data.html}
\BIBentrySTDinterwordspacing

\bibitem{lightcastle_partner}
\BIBentryALTinterwordspacing
``Bangladesh rice industry: Essential for rural development.'' [Online]. Available: \url{https://www.lightcastlebd.com/insights/2019/10/rice-industry-bangladesh/}
\BIBentrySTDinterwordspacing

\bibitem{10273722}
N.~Bharanidharan, S.~R.~S. Chakravarthy, H.~Rajaguru, V.~V. Kumar, T.~R. Mahesh, and S.~Guluwadi, ``Multiclass paddy disease detection using filter-based feature transformation technique,'' \emph{IEEE Access}, vol.~11, pp. 109\,477--109\,487, 2023.

\bibitem{AHAD202322}
\BIBentryALTinterwordspacing
M.~T. Ahad, Y.~Li, B.~Song, and T.~Bhuiyan, ``Comparison of cnn-based deep learning architectures for rice diseases classification,'' \emph{Artificial Intelligence in Agriculture}, vol.~9, pp. 22--35, 2023. [Online]. Available: \url{https://www.sciencedirect.com/science/article/pii/S2589721723000235}
\BIBentrySTDinterwordspacing

\bibitem{SINGH20232026}
\BIBentryALTinterwordspacing
S.~P. Singh, K.~Pritamdas, K.~J. Devi, and S.~D. Devi, ``Custom convolutional neural network for detection and classification of rice plant diseases,'' \emph{Procedia Computer Science}, vol. 218, pp. 2026--2040, 2023, international Conference on Machine Learning and Data Engineering. [Online]. Available: \url{https://www.sciencedirect.com/science/article/pii/S1877050923001795}
\BIBentrySTDinterwordspacing

\bibitem{Haridasan2023}
\BIBentryALTinterwordspacing
A.~Haridasan, J.~Thomas, and E.~D. Raj, ``Deep learning system for paddy plant disease detection and classification,'' \emph{Environmental Monitoring and Assessment}, vol. 195, pp. 1--28, 1 2023. [Online]. Available: \url{https://link.springer.com/article/10.1007/s10661-022-10656-x}
\BIBentrySTDinterwordspacing

\bibitem{KM2023105836}
\BIBentryALTinterwordspacing
S.~K.M., S.~V., S.~K. P., and S.~O.K., ``Ai based rice leaf disease identification enhanced by dynamic mode decomposition,'' \emph{Engineering Applications of Artificial Intelligence}, vol. 120, p. 105836, 2023. [Online]. Available: \url{https://www.sciencedirect.com/science/article/pii/S0952197623000209}
\BIBentrySTDinterwordspacing

\bibitem{su151612149}
\BIBentryALTinterwordspacing
M.~Aggarwal, V.~Khullar, N.~Goyal, A.~Alammari, M.~A. Albahar, and A.~Singh, ``Lightweight federated learning for rice leaf disease classification using non independent and identically distributed images,'' \emph{Sustainability}, vol.~15, no.~16, 2023. [Online]. Available: \url{https://www.mdpi.com/2071-1050/15/16/12149}
\BIBentrySTDinterwordspacing

\bibitem{agronomy13040961}
\BIBentryALTinterwordspacing
C.~G. Simhadri and H.~K. Kondaveeti, ``Automatic recognition of rice leaf diseases using transfer learning,'' \emph{Agronomy}, vol.~13, no.~4, 2023. [Online]. Available: \url{https://www.mdpi.com/2073-4395/13/4/961}
\BIBentrySTDinterwordspacing

\bibitem{Aggarwal2022}
S.~Aggarwal, M.~Suchithra, N.~Chandramouli, M.~Sarada, A.~Verma, D.~Vetrithangam, B.~Pant, and B.~A. Adugna, ``Rice disease detection using artificial intelligence and machine learning techniques to improvise agro-business,'' \emph{Scientific Programming}, vol. 2022, 2022.

\bibitem{Prottasha_et}
M.~S.~I. Prottasha, A.~Hossain, M.~Rahman, S.~M.~S. Reza, and D.~Hossain, ``Identification of various rice plant diseases using optimized convolutional neural network,'' \emph{IJCDS Journal}, vol.~12, pp. 1539--1551, 12 2022.

\bibitem{Islam2022}
\BIBentryALTinterwordspacing
M.~N. Islam, F.~Ahmed, M.~T. Ahammed, M.~Rashid, and B.~S. Bari, ``Rice disease identification through leaf image and iot based smart rice field monitoring system,'' \emph{Lecture Notes in Electrical Engineering}, vol. 900, pp. 529--539, 2022. [Online]. Available: \url{https://link.springer.com/chapter/10.1007/978-981-19-2095-0_45}
\BIBentrySTDinterwordspacing

\bibitem{Islam2021}
\BIBentryALTinterwordspacing
A.~Islam, R.~Islam, S.~M.~R. Haque, S.~M.~M. Islam, M.~Ashik, and I.~Khan, ``Intelligent systems and applications,'' \emph{Intelligent Systems and Applications}, vol.~5, pp. 35--45, 2021. [Online]. Available: \url{http://www.mecs-press.org/}
\BIBentrySTDinterwordspacing

\bibitem{Hossain2021}
M.~F. Hossain, S.~Abujar, S.~R.~H. Noori, and S.~A. Hossain, ``Dhan-shomadhan: A dataset of rice leaf disease classification for bangladeshi local rice,'' vol.~1, 2021.

\bibitem{Rice_Leafs}
\BIBentryALTinterwordspacing
S.~RIYAZ, ``Rice leafs.'' [Online]. Available: \url{https://www.kaggle.com/datasets/shayanriyaz/riceleafs}
\BIBentrySTDinterwordspacing

\bibitem{dalal2005histograms}
\BIBentryALTinterwordspacing
N.~Dalal and B.~Triggs, ``Histograms of oriented gradients for human detection.'' [Online]. Available: \url{https://lear.inrialpes.fr/people/triggs/pubs/Dalal-cvpr05.pdf}
\BIBentrySTDinterwordspacing

\bibitem{ojala2002multiresolution}
T.~Ojala, M.~Pietikainen, and T.~Maenpaa, ``Multiresolution gray-scale and rotation invariant texture classification with local binary patterns,'' \emph{IEEE Transactions on Pattern Analysis and Machine Intelligence}, vol.~24, no.~7, pp. 971--987, 2002.

\bibitem{resnet50}
\BIBentryALTinterwordspacing
K.~He, X.~Zhang, S.~Ren, and J.~Sun, ``Deep residual learning for image recognition.'' [Online]. Available: \url{https://www.cv-foundation.org/openaccess/content_cvpr_2016/papers/He_Deep_Residual_Learning_CVPR_2016_paper.pdf}
\BIBentrySTDinterwordspacing

\bibitem{inception}
C.~Szegedy, V.~Vanhoucke, S.~Ioffe, and J.~Shlens, ``Rethinking the inception architecture for computer vision.''

\bibitem{simonyan2015deep}
K.~Simonyan and A.~Zisserman, ``Very deep convolutional networks for large-scale image recognition,'' 2015.

\bibitem{48080}
A.~Howard, A.~Zhmoginov, L.-C. Chen, M.~Sandler, and M.~Zhu, ``Inverted residuals and linear bottlenecks: Mobile networks for classification, detection and segmentation,'' in \emph{CVPR}, 2018.

\bibitem{pmlr-v97-tan19a}
\BIBentryALTinterwordspacing
M.~Tan and Q.~Le, ``{E}fficient{N}et: Rethinking model scaling for convolutional neural networks,'' in \emph{Proceedings of the 36th International Conference on Machine Learning}, ser. Proceedings of Machine Learning Research, K.~Chaudhuri and R.~Salakhutdinov, Eds., vol.~97.\hskip 1em plus 0.5em minus 0.4em\relax PMLR, 09--15 Jun 2019, pp. 6105--6114. [Online]. Available: \url{https://proceedings.mlr.press/v97/tan19a.html}
\BIBentrySTDinterwordspacing

\bibitem{8237336}
R.~R. Selvaraju, M.~Cogswell, A.~Das, R.~Vedantam, D.~Parikh, and D.~Batra, ``Grad-cam: Visual explanations from deep networks via gradient-based localization,'' in \emph{2017 IEEE International Conference on Computer Vision (ICCV)}, 2017, pp. 618--626.

\end{thebibliography}

\vspace{12pt}

\end{document}